\documentclass[conference,compsoc]{IEEEtran}
\ifCLASSOPTIONcompsoc
  \usepackage[nocompress]{cite}
\else
  \usepackage{cite}
\fi
\ifCLASSINFOpdf
\else
\fi
\usepackage{graphicx}
\usepackage{hyperref}
\usepackage{xcolor}
\usepackage{comment}

\usepackage{graphicx}
\usepackage{comment}

\begin{document}
%
\title{AutoML-Med: a tool for optimizing pipeline generation in medical ML}

\author{\IEEEauthorblockN{Riccardo Francia, Giorgio Leonardi, \\Stefania Montani, Marzio Pennisi, \\Manuel Striani}
\IEEEauthorblockA{DISIT\\Computer Science Institute\\
University of Piemonte Orientale\\
Alessandria, Italy\\
Email: name.surname@uniupo.it}
\and
\IEEEauthorblockN{Maurizio Leone}
\IEEEauthorblockA{Fondazione IRCCS\\Casa Sollievo della Sofferenza\\San Giovanni Rotondo\\
Foggia, Italy\\
Email: m.leone@operapadrepio.it}
\and
\IEEEauthorblockN{Sandra D'Alfonso}
\IEEEauthorblockA{DISS\\
University of Piemonte Orientale\\
Novara, Italy\\
Email: sandra.dalfonso@med.uniupo.it}
}

%


\maketitle

\begin{abstract}
Medical datasets are typically affected by issues such as missing values, class imbalance, a
heterogeneous feature types, and a high number of features versus a relatively small
number of samples, preventing machine learning models from obtaining proper results in classification and regression tasks. This paper introduces AutoML-Med, an Automated Machine Learning tool specifically designed to address these challenges, minimizing user intervention and identifying the optimal combination of preprocessing techniques and predictive models. AutoML-Med's architecture incorporates Latin Hypercube Sampling (LHS) for exploring preprocessing methods, trains models using selected metrics, and utilizes Partial Rank Correlation Coefficient (PRCC) for fine-tuned optimization of the most influential preprocessing steps. Experimental results demonstrate AutoML-Med's effectiveness in two different clinical settings, achieving higher balanced accuracy and sensitivity, which are crucial for identifying at-risk patients, compared to other state-of-the-art tools. AutoML-Med's ability to improve prediction results, especially in medical datasets with sparse data and class imbalance, highlights its potential to streamline Machine Learning applications in healthcare.
\end{abstract}


%
\IEEEpeerreviewmaketitle

\section{Introduction}
In recent years, the advent of deep learning and, in particular, transformer-based architectures, has significantly revolutionized the field of Artificial Intelligence (AI) in many scientific domains, including computer vision, natural language processing, and sequence modeling, thanks to the increasing availability of computational power and large-scale data-sets. However, classical Machine Learning (ML) methods, such as decision trees, gradient-boosted trees, Support Vector Machines (SVMs), and regression—based techniques, continue to be considered as the state-of-the-art for tabular data, which are still nowadays widely used in healthcare, finance, industrial monitoring, and other structured-data domains. There are several reasons for this. Notably, conventional AI models tend to perform reasonably well on datasets of limited size, whereas state-of-the-art deep learning techniques typically require substantially larger amounts of data to generalize effectively. Moreover, many classical AI methods, such as regression, Bayesian approaches, rule-based systems, and tree-based models, are inherently more interpretable, a characteristic that is particularly valuable in high-stakes domains such as healthcare. In contrast, deep learning models often work as black boxes, limiting their explainability.

As an example, Grinsztajn et al. \cite{Grinsztajn2022} showed that tree-based ensembles like XGBoost and Random Forests consistently outperformed a wide range of contemporary deep learning models across dozens of medium-sized tabular datasets ($\approx$ 10.000 samples), even when an extensive hyperparameter optimization was taken into account. This performance gap is primarily attributed to the inductive biases of tree models: they are inherently robust to uninformative features, adept at capturing irregular (non-smooth) decision boundaries, and unaffected by rotational transformations of the input space, which are challenging for neural networks \cite{Grinsztajn2022, GrinsztajnNeurIPS2022}.

Despite ongoing research into transformer- and neural-specific architectures for tabular data, such as TabTransformer, TabNet, SAINT, and FT-Transformer, the most recent comprehensive surveys and controlled studies conclude that classical ML approaches, and in particular ensemble-based tree methods, remain the dominant solution for tabular-structured problems \cite{GarciaSurvey2021, Grinsztajn2022}.

However, real-world tabular datasets, particularly in sensitive domains such as medicine, often present additional challenges such as missing values, imbalanced class distributions, noisy measurements,  heterogeneous feature types, and a high number
of features versus a relatively small number of samples. These issues can significantly impair model performance if not addressed properly. Consequently, the data preprocessing stage becomes a critical step in the ML pipeline, often having a greater impact on final performance than model selection itself \cite{kuhn2019feature, garcia2015preprocessing}. Techniques such as imputation, normalization, feature selection, resampling strategies (e.g., SMOTE), and encoding schemes are essential for enhancing the quality and usability of the data. Particularly in the medical domain, where data sparsity and class imbalance are common, carefully designed preprocessing workflows can substantially improve both model robustness and interpretability \cite{rahman2013handling, chandrashekar2014feature}.

 In this work, we present an Automated Machine Learning (AutoML) tool, named AutoML-Med,  specifically designed to handle the typical challenges encountered in medical tabular datasets. We have applied this tool to a clinical database developed to monitor disease progression in patients with Multiple Sclerosis (MS), a chronic inflammatory demyelinating disease of the central nervous system  that primarily affects young adults, leading to severe physical disability. The dataset exhibits several common but critical issues, including missing values, class imbalance, redundant or irrelevant features, and inconsistent entries.
To maximize the extraction of meaningful information and support the development of robust predictive models for disease progression, the AutoML-Med framework incorporates semi-automated selection of pre-processing steps, automated feature selection, and tailored strategies for handling imbalanced and incomplete data. Importantly, the tool requires minimal user intervention and no advanced programming skills, making it particularly suitable for clinicians and researchers without a technical background. 
We show the results obtained with our tool in comparison with other state-of-the-art tools in the field, demonstrating solid performance. 

Furthermore, since the MS dataset is not publicly available due to privacy and ethical constraints, we have assessed the reproducibility of our approach by applying the AutoML-Med tool to a publicly available dataset on Type 2
Diabetes risk prediction. We have positively compared the results obtained with our tool against previously published benchmarks on the same dataset.

The paper is organized as follows: section \ref{related} presents related work; section \ref{tool} provides the details of AutoML-Med; section \ref{results} showcases our experimental results, and section \ref{conclu} is devoted to conclusions.

\section{Related Work}
\label{related}



In a typical ML pipeline, practitioners begin with a collection of input data points for training purposes. However, this raw data is often not in a format suitable for direct use by all algorithms. To prepare the data for effective modeling, experts may need to apply various techniques such as data preprocessing, feature engineering, feature extraction, and feature selection. Once the data is properly prepared, the next steps involve selecting an appropriate algorithm and tuning its hyperparameters to achieve optimal predictive performance. Each of these stages can be complex and time-consuming, posing significant challenges for those working with ML.

Automated Machine Learning (AutoML) is an area of research that emerged after 2010, aiming to streamline this process by automating the iterative and labor-intensive tasks involved in building and optimizing a ML pipeline for deployment. 



Several AutoML tools are nowdays available to researchers. 
Auto-sklearn \cite{Feurer2019}, which is built upon the scikit-learn library \cite{DBLP:journals/corr/abs-1201-0490}, employs a Bayesian optimization algorithm to automatically select a pipeline that includes a data preprocessing step, a feature selection method, and a ML classifier. It was the first AutoML framework to incorporate both meta-learning and ensemble classification to enhance pipeline performance. Notably, Auto-sklearn leverages meta-data derived from the analysis of numerous public datasets to guide the optimization of pipelines for new tasks.


H2O \cite{ledell} is a ML framework, similar to scikit-learn, that offers a suite of algorithms capable of running on a server cluster. This framework can be accessed through various interfaces and programming languages. H2O includes an AutoML module that builds ML pipelines using its proprietary algorithms. Configuration options are relatively minimal, limited to selecting algorithms, defining stopping criteria, and specifying the extent of k-fold cross-validation. The system conducts an exhaustive search across its feature engineering strategies and model hyperparameters to optimize pipeline performance.

The General Automated Machine learning Assistant (GAMA) \cite{DBLP:conf/pkdd/GijsbersV20} is a modular AutoML platform designed to give users more transparency and control over the pipeline search process. Unlike many current AutoML systems that operate as black boxes, GAMA enables users to integrate different AutoML components and post-processing methods, log and visualize the optimization process, and conduct benchmarking. It  includes three AutoML search algorithms, two post-processing techniques, and is built to support the integration of additional components. It also features a version for imbalanced datasets. 

The Tree-based Pipeline Optimization Tool (TPOT) \cite{DBLP:conf/gecco/OlsonBUM16} introduces the use of evolutionary algorithms to automate end-to-end ML pipelines. TPOT represents pipelines as expression trees and employs genetic programming to discover and optimize them. This approach assembles preprocessing and modeling operators into compact tree structures that outperform basic baselines, guided by a Pareto optimization that balances accuracy and complexity. Later methodological reviews identified TPOT as one of the earliest and most widely adopted AutoML “innovation engines,” built using the scikit-learn Python library \cite{DBLP:journals/corr/abs-1201-0490}.

A more recent development, described in \cite{DBLP:conf/gptp/RibeiroSMMCHM23}, is TPOT2—a complete reimplementation of TPOT that replaces its tree-based pipeline representation with a directed acyclic graph (DAG) structure called "GraphPipelineIndividual". This transition allows each step in the pipeline, such as data cleaning, feature transformation, or model stacking, to be represented explicitly as a node, removing TPOT's rigid “one-root, many-leaf” structure. As a result, the new system supports arbitrary branching, bypassing, and recursive sub-pipelines. TPOT2’s evolutionary engine was redesigned from the ground up, modularizing  components such as parent selection, mutation (with eight operators), and crossover (with two operators), which are now pluggable. The entire search space—including root, inner, and leaf nodes—is defined using simple Python dictionaries that integrate seamlessly with Optuna-compatible samplers, enabling joint evolutionary and Bayesian hyperparameter optimization. 


As mentioned above, a wide range of AutoML frameworks exist, which can  deliver  effective results with minimal user intervention. While these frameworks often rely on a common set of standardized techniques, the strategies used to automate their application and evaluation can vary significantly. Several studies have addressed the challenge of benchmarking by comparing different AutoML systems on the same datasets (see, for example, \cite{DBLP:journals/corr/abs-1808-06492} and \cite{DBLP:journals/corr/abs-2211-00376}, the latter focusing specifically on imbalanced datasets). Additionally, the work in \cite{DBLP:journals/biodatamining/RomeroDMJSMM22} benchmarks the application of various AutoML tools in the context of clinical data.


Indeed, the use of ML techniques has shown clear potential to enhance health outcomes, reduce healthcare costs, and accelerate clinical research. Despite this promise, most hospitals have yet to implement reliable  ML solutions. A key barrier is that healthcare professionals often lack the specialized ML expertise needed to develop effective models and integrate them seamlessly into clinical workflows. AutoML offers a valuable way to address these challenges. However, healthcare datasets pose unique difficulties for AutoML systems, such as large sample sizes, pronounced class imbalance, and many missing data.

The survey in \cite{DBLP:journals/artmed/WaringLU20}, which reviewed over 100 studies in AutoML on healthcare, has found that these automated methods can improve performance in several ML tasks, often completing them more quickly. Nevertheless, a major current limitation of AutoML is scaling effectively beyond small- and medium-sized retrospective datasets.

The previously mentioned study in \cite{DBLP:journals/biodatamining/RomeroDMJSMM22} benchmarks three AutoML tools (Auto-sklearn \cite{Feurer2019}, H2O \cite{ledell}, and TPOT \cite{DBLP:conf/gecco/OlsonBUM16}) on clinical data, reporting only modest outcomes.

These limitations justify, especially in the medical domain, the study of novel approaches, like the one we are presenting in this paper.


\section{The AutoML-Med tool}
\label{tool}


AutoML-Med is an approach aiming to identify the best combination of preprocessing techniques and predictive models for classification and regression, minimizing human intervention and making use of established best practices. In particular, we target medical tabular datasets, characterized by missing data, strong class imbalance and a high number of features versus a relatively small number of samples. 


Our tool implements an automated pipeline generation  selecting an optimal combination of preprocessing techniques and predictive models through the use of stratified sampling and sensitivity analysis.
Its main steps are shown in figure \ref{fig:archi}, and are illustrated in the following subsections.

\subsection{Preprocessing methods sampling}

As regards the preprocessing phase, our pipeline is structured in five ordered and sequential steps: 1) Missing data imputation; 2) Class balancing; 3) Feature engineering; 4) Feature scaling; 5) Feature selection. 
Each preprocessing step is represented as a list of possible techniques, as shown in the upper-left corner of the architecture in Figure \ref{fig:archi}. Even if we primarily target medical imbalanced datasets, the methods we propose have been selected from the scikit-learn and imbalanced-learn libraries and are intended for general use; the user can extend the list with other methods using the scikit-learn API. This flexibility allows the pipeline to be adapted to the nature of the dataset and the purpose of the analysis.
The pipeline space is thus defined as the Cartesian product of the lists of preprocessing techniques for each of the five steps. Each possible pipeline will be a unique combination of preprocessing methods, and AutoML-Med aims at suggesting the best one.

To reduce the computational complexity, 
we borrowed from the world of stochastic sensitivity analysis
a technique named Latin Hypercube Sampling - Partial Rank Correlation Coefficient
(LHS-PRCC), which is a statistical analysis technique extensively utilized in computational modeling to understand which input parameters are fundamental and most influence the variability of a model’s outcome \cite{marino2008}. This methodology was developed to overcome the limitations of classical sensitivity
analysis, where results are often strongly dependent on the specific values of fixed parameters.
The LHS-PRCC approach integrates two main components:
\begin{itemize}
    \item {\bf Latin Hypercube Sampling (LHS)} \cite{McKay1979}: This technique allows to sample the entire input parameter space in a stratified manner, by dividing each random parameter distribution into probability intervals (usually with normal or uniform distributions), from which samples are then drawn. This method allows to test the simultaneous variation of all parameters, thus avoiding the limitations of fixing other parameters.
\item {\bf Partial Rank Correlation Coefficient (PRCC)} \cite{Saltelli1990}: After the LHS procedure generates a unique set of variables for the model, and the model is solved for each set, the PRCC then computes a partial correlation on rank-transformed
data between the input parameters (model parameters) and the output values (entities’ behaviors). PRCC values range from -1 to 1, with values near 1 indicating a strong positive correlation, and values near -1 indicating a strong
negative correlation; clearly, values near 0  suggest no or little correlation between input and output values.
\end{itemize}
A significant advantage of LHS-PRCC is that the correlation indexes obtained do not depend on a given set of fixed parameters, allowing the estimation of a parameter’s influence regardless of the values of other parameters.

In a nutshell, the idea is to apply this technique to identify which steps and parameters of the pre-processing pipeline (i.e., the input variables) have the greatest impact on the performance of the ML model, in terms of metrics such as accuracy, specificity, etc.
Specifically, we first employ LHS, due to its efficiency in exploring high-dimensional parameter spaces, to generate a sub-optimal set of pre-processing parameter combinations. We then apply PRCC analysis to determine the most influential parameters i.e., those most strongly correlated with model performance, on which to focus subsequent fine-tuning.

LHS (phase (a) of the architecture in Figure \ref{fig:archi}) operates by splitting the space of choices into equiprobable intervals to ensure that each interval is represented at least once. This approach enables us to have a more uniform coverage of the space of configurations than using random sampling. Since LHS operates natively on continuous variables, we had to adapt the problem to use stratified sampling on a discrete space, as follows: for each preprocessing step $s$ with $N_s$ possible methods, we define $N_s$ intervals. For each pipeline, LHS samples a value $x_s \in [0,N_s)$ for each step $s$. The sampled value is then discretized by taking its integer part and is used as an index to select the corresponding preprocessing method for step $s$.

After the sampling phase, each pipeline is applied to the  dataset, and a preprocessed dataset (with data imputed and  classes balanced by means of specific techniques) is generated. This phase is reported as phase (b) of the architecture in Figure \ref{fig:archi}.

The tool presents the possibility of being customized: the users can specify the preprocessing methods to be considered, they can skip one or more preprocessing steps, and, if necessary, they can choose to operate on a sampled subset of the original dataset to further speed up the process.

\subsection{Prediction model training and selection}

\begin{figure*}[htbp]
\begin{center}
\includegraphics[width=0.8\textwidth]{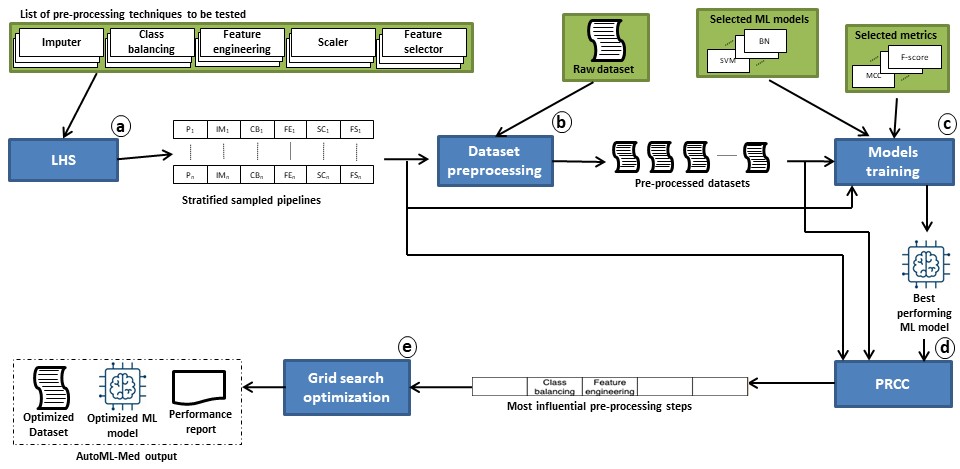}
\end{center}
\caption{System's architecture. Green squares contain input data; blue squares the AutoML-Med processing phases, whose sequence is indicated by the circled letters.}\label{fig:archi}
\end{figure*}

On every preprocessed dataset, one or more ML model is trained. This task is accomplished by the phase (c) of the architecture in Figure \ref{fig:archi}. The evaluation process is parallelized to reduce execution time. During the evaluation phase, to reduce the risk of overfitting, a random cross-validation search is performed for the best hyperparameters for the predictive model being trained. Given the imbalanced nature of the data, the set of evaluation metrics that are calculated to find the optimal pipeline  includes balanced accuracy, $F1-score$, $F\beta-score$ (with $\beta=0.5$), and Matthews Correlation Coefficient (MCC).

A table is constructed to represent the structure of the sampled pipelines and the value of the target metrics calculated on the trained predictive models. The table has as many rows as the sampled pipelines, five columns representing the preprocessing methods that compose a given pipeline, and as many columns as the calculated metrics obtained for each of the models being tested.
At the end of the evaluation phase, the predictive model that yielded the best overall result on the  target metric(s) is isolated, and then the other models are discarded.

The tool presents the possibility of being customized in this phase as well: the users can specify the ML models they want to include in the evaluation and the target metric(s) to be optimized. It is also possible to set a global seed to ensure the reproducibility of the experiments.

\subsection{Fine-tuned optimization}



Once the best performing prediction model has been selected, on the basis of the specified target metric(s), AutoML-Med allows to fine tune the preprocessing pipeline choices. Indeed, since the preprocessing methods were selected through a sampling procedure, it is possible that optimal choices were excluded. Once the predictive model has been locked, a finer approach can be applied.

In particular, AutoML-Med identifies the most impacting preprocessing phases, enabling the search to focus just on them. To this end, it exploits sensitivity analysis adopting the Partial Rank Correlation Coefficient (PRCC) \cite{gasior_examining_2024}. PRCC, executed in the phase (d) of the architecture in Figure \ref{fig:archi},  differs from traditional correlation measures in that it allows the effect of each input variable to be isolated, removing the influence of the remaining ones.


To adopt PRCC, the preprocessing steps are defined as the independent variables. The dependent variable (or output) is represented by the value of the target metric(s) obtained from a given predictive model.
To make the correlation calculation more robust, it is necessary to perform a ranking operation on the independent variables. Assigning each value its own rank within the distribution neutralizes the differences given by the scales of the independent variables.\\

Let Y be the dependent variable and let $\{X_1, X_2,X_3, \cdots , X_k \}$ be the $k$ independent variables; the process to calculate the PRCC for each independent variable is as follows:

\begin{enumerate}
    \item The regression of $X_i$ against the remaining input variables $\{X_j\}_{j\neq i}$ is performed. The residual $\epsilon_i$ of the regression represents the component of $X_i$ not explained by the other independent variables.

    \item A regression of the output variable $Y$ is run against the same variables $\{X_j\}_{j\neq i}$. The residual $\delta_i$ of the regression represents the component of $Y$ not explained by the other independent variables.

    \item The $PRCC$ value for the variable $X_i$ is calculated as $PRCC(X_i,Y)=corr(\epsilon_i , \delta_i)$
\end{enumerate}

The procedure is repeated for each independent variable, a high absolute value indicates a direct and independent impact on output.


Our tool thus proceeds by taking the $m$ most influential preprocessing steps as identified by PRCC ($m$=2 by default), and creates $n$ variations of them through the use of a grid search, while keeping the other steps unchanged.

This strategy, implemented as the final phase (e) in Figure \ref{fig:archi}, allows for in-depth exploration of combinations of preprocessing techniques on the steps that demonstrated a large impact on model performance, thus increasing the possibility of identifying better configurations that may not have been included in the initial stratified sampling step.

The evaluation of the new pipelines is carried out again in parallelized mode, using only the model that maximized the target metric(s) in the previous step. At the end of this phase, the pipeline that obtains the highest value on the target metric(s) is selected and presented along with a complete report of its performance.

\section{Experimental Results}
\label{results}

In this section, we present our experimental results, conducted in two different medical domains. The first is the one of Multiple Sclerosis: to this end, we worked on a proprietary dataset, available in the context of the EU project WISDOM\footnote{https://emsp.org/projects/wisdom/}. The second application refers to Type 2 Diabetes, and resorts to a publicly available dataset. In this case, we could compare our approach to the one published in \cite{xie2019building}, which operated on the same data.
Details are provided in the following subsections.

\subsection{Multiple Sclerosis risk prediction}
Multiple Sclerosis (MS) is a chronic inflammatory demyelinating disorder of the central nervous system  that primarily affects young adults and can eventually lead to severe physical disability \cite{Attfield2022}. The disease is characterized by the
loss of myelin sheath from axons of the brain
and spinal cord, which results in reduced communication
among nerve cells. This damage disrupts the ability of the nervous system to communicate, leading to a wide range of neurological signs and symptoms.

The etiology of MS is only partially understood, and the response to treatment is highly variable. Autoimmunity plays a central role in the disease pathogenesis, as the body’s own immune system attacks the myelin sheath, leading to demyelination and neurodegeneration \cite{Attfield2022}. Several risk factors have been implicated in the development of MS, including genetic susceptibility \cite{IMSGC-xo}, environmental influences such as Epstein-Barr virus (EBV) infection, low vitamin D levels, smoking, and dietary components, as well as epigenetic modifications \cite{Alfredsson2019-vf}.

Globally, about 2.3 to 2.5 million people live with MS, with prevalence rates ranging between 2 and 150 per 100,000 individuals depending on geographic and population-specific factors \cite{Walton2020}. In Europe, MS is the leading cause of non-traumatic neurological disability in young adults, affecting over 700,000 individuals \cite{Walton2020}.

Relapsing-Remitting Multiple Sclerosis (RRMS) is the most prevalent clinical form, accounting for approximately 85–90\% of all MS cases at onset. RRMS is characterized by episodes of neurological dysfunction (relapses), followed by periods of partial or complete recovery (remissions), with varying degrees of cumulative disability over time \cite{Attfield2022}.

In this context, we analyzed a medical tabular dataset that was used to collect many information about MS disease onset and progression in Italy, including features such as age at onset, sex, genetic factors, presence of oligoclonal bands, MS form,  disability status at the onset measured according to the Extended Disability Status Scale (EDSS), etc.
To evaluate our framework, we used this medical dataset that presented most of the main issues we aimed to address when defining AutoML-Med: many missing data, a strong class imbalance and a high number of features versus a  small number of samples. 
In particular, the dataset consisted of 1,031 real patients and 33 features with a positive/negative ratio of 0.27 The features contained in the dataset describe: demographic and geographic data, clinical and disease history data, genetic data, disease familiarity, neurological functions and involvement in initial symptoms, sequelae and indicators of complications. The content of the
dataset is confidential and was granted in the context of the
PROGEMUS and WISDOM project. 
The model was trained with the aim of classifying a patient's individual risk of reaching disability level 4, modelled on the EDSS scale.

To benchmark the results obtained by AutoML-Med, we chose three other frameworks that allow for auto-ML: Auto-sklearn \cite{Feurer2019}, GAMA \cite{DBLP:conf/pkdd/GijsbersV20}, and AutoBalance, which is a fork of GAMA optimized for imbalanced datasets\footnote{https://arxiv.org/pdf/2211.00376}.

For all the tested frameworks, we performed 10 independent runs with varying seeds. Each run involves a stratified split of the dataset in order to allocate 2/3 of the data to the training set and 1/3 of the data to the test set. Table~\ref{tab:wisdom} shows the metrics of interest calculated as the average obtained from the 10 runs performed. For each metric, the standard deviation relative to the values collected is also reported.

\begin{table}[ht]
    \centering
    \resizebox{\linewidth}{!}{
        \begin{tabular}{|l|c|c|c|c|c|c|}
            \hline
            Framework & Balanced Accuracy & Sensitivity & Specificity & F-1 macro & MCC & AUC\\
            \hline
            Auto-sklearn        & 0.8472 & 0.7158 & 0.9786 & 0,8735 & 0,7590 & 0,9187 \\
            GAMA                & 0.8492 & 0.7158 & 0.9826 & 0.8787 & 0.7683 & 0.9100 \\
            AutoBalance         & 0.8466 & 0.7492 & 0.944  & 0.8534 & 0.7111 & 0.9130 \\
            \hline
            AutoML-Med          & 0.8896 & 0.8539 & 0.9253 & 0.8745 & 0.7558 & 0.8896 \\
            \hline
        \end{tabular}
    }
    \caption{Comparison between metrics in different models on MS patients} 
    \label{tab:wisdom}
\end{table}

Among the results in Table~\ref{tab:wisdom}, we firstly focus on balanced accuracy, which is particularly suitable to deal with imbalanced datasets. AutoML-Med achieved the highest balanced accuracy (0.8896 ±0.0201) among all the tested  frameworks, demonstrating a superior ability to correctly classify patients belonging to both the positive and the negative classes. Another important factor is the ability of the classifier to reduce false negatives, in order to prevent that potential patients, if predicted as negative, miss a proper treatment. In this direction, Table~\ref{tab:wisdom} shows that AutoML-Med achieved the highest sensitivity (0.8539 ±0.0500) of the pool, demonstrating an appreciable advantage in identifying patients at risk (positive class). On the other side, AutoML-Med has lower specificity values than some competing frameworks. This means that more false positives are predicted, but considering the balance between sensitivity and specificity, and given the importance to correctly identify patients at risk, this can be considered acceptable in this clinical context.
The F1 macro and MCC metrics obtained by AutoML-Med are in line with the best competing frameworks, thus confirming the robustness of its overall performance.



\subsection{Type 2 Diabetes risk prediction}
We tested our tool also using the publicly available Behavioral Risk Factor Surveillance System (BRFSS) 2014 dataset\footnote{https://www.cdc.gov/brfss/annual\_data/annual\_2014.html}, which contains preventive health practices and risk behaviors that are linked to chronic diseases, injuries, and preventable infectious diseases that affect the adult population in the US. We compared the obtained metrics  with the results reported in 
\cite{xie2019building}, which contains an analysis about risk factors for Type 2 Diabetes, a chronic disease that increases risk for stroke, kidney failure, renal complications, peripheral vascular disease, heart disease, and death \cite{diab-on}. 
The BRFSS dataset consists of 464,664 samples and 279 variables. 

To allow a fair comparison, we applied the data filtering procedure proposed by \cite{xie2019building}, but without removing patients with missing data, as our framework is able to handle missing data through imputation. At the end of the filtering phase, the resulting dataset had 27 variables, 76,156 patients without Type 2 Diabetes (negative class) and 14,532 patients with Type 2 Diabetes (positive class), with a positive/negative ratio of 0.19.
To run AutoML-Med, we chose to use the three predictive models proposed by \cite{xie2019building} that achieved the best results in terms of accuracy and ROC: Logistic Regression, Support Vector Machine, and Neural Network. We adopted a stratified split, assigning 2/3 of the data to the training set and 1/3 to the test set.  Table ~\ref{tab:modelli} shows a comparison between the metrics presented by \cite{xie2019building} and the average of the metrics obtained by running ten independent runs in AutoML-Med, each with a different seed, in order to ensure a more reliable estimate of the performance.

\begin{table}[ht]
    \centering
    \resizebox{\linewidth}{!}{
        \begin{tabular}{|l|c|c|c|c|}
            \hline
            Model & Balanced Accuracy & Sensitivity & Specificity & AUC \\
            \hline
            Neural Network      & 0.6398 & 0.3781 & 0.9016 & 0.7949 \\
            Logistic Regression & 0.665  & 0.4634 & 0.8666 & 0.7932 \\
            Linear SVM          & 0.6503 & 0.4260 & 0.8746 & 0.7807 \\
            RBF SVM             & 0.6458 & 0.4014 & 0.8902 & 0.7788 \\
            \hline
            AutoML-Med       & 0.7436 & 0.7968 & 0.6904 & 0.7436 \\
            \hline
        \end{tabular}
    }
    \caption{Comparison between metrics in different models on Type 2 Diabetes patients}
    \label{tab:modelli}
\end{table}

The paper in\cite{xie2019building} 
does not report the balanced accuracy value, so we decided to calculate it retrospectively using the formula $(Sensitivity + Specificity) / 2 $. In the context of imbalanced datasets, using balanced accuracy to evaluate the performance of a classifier is preferable to using accuracy, as the latter can be optimistic or misleading, rewarding a model that accurately identifies only the majority class.
All the experiments reported in \cite{xie2019building}
report very high specificity and very low sensitivity. This means that the predictive models reported correctly identify most patients who are not at risk (negative class, which is the majority), but may not recognize many patients who are at risk. The high AUC is probably influenced by the ability to correctly classify the majority class, as is often the case in imbalanced datasets.

Our tool, on the other hand, leads to a much higher sensitivity, correctly identifying most patients at risk. The specificity is lower, so the number of false positives increases, but this is an acceptable compromise in the medical field, where it is preferable not to miss patients at risk. The balanced accuracy is higher, indicating better overall performance across the two classes.

\section{Conclusion}
\label{conclu}

This paper introduced AutoML-Med, an Auto-ML tool designed to address challenges in medical tabular datasets, such as missing values, class imbalance, and heterogeneous feature types. AutoML-Med minimizes user intervention and aims to identify the best combination of preprocessing techniques and predictive models for classification and regression tasks. The tool's architecture incorporates Latin Hypercube Sampling (LHS) for approximate preprocessing method exploration, model training with selected metrics as output, and Partial Rank Correlation Coefficient (PRCC) for fine-tuned optimization of the most influential preprocessing steps.

Experimental results compared AutoML-Med's performance in different clinical settings. In Multiple Sclerosis risk prediction,  our framework achieved higher balanced accuracy and sensitivity. In Type 2 Diabetes risk prediction, AutoML-Med also showed significantly higher sensitivity and balanced accuracy, crucial for identifying at-risk patients. The tool's ability to enhance the prediction results, particularly in data-sparse and class-imbalanced medical datasets, underscores its potential to streamline ML applications in healthcare.

\vspace{1em}

To support reproducibility and maintain compliance with double-blind review, the code will be shared with interested researchers after the review process.


\ifCLASSOPTIONcompsoc
  \section*{Acknowledgments}
\else
  \section*{Acknowledgment}
\fi

This work is partially supported by the project WISDOM (Well-being improvement through the Integration of healthcare and research Data and models with Out border for chronic iMmune-mediated diseases). WISDOM has received funding from the European Union’s Horizon Europe Research and Innovation Actions under grant no. 101137154.



\bibliographystyle{IEEEtran}
\bibliography{biblio}
%

\end{document}